\documentclass[letterpaper]{article} % DO NOT CHANGE THIS
\usepackage{aaai25}  % DO NOT CHANGE THIS  [submission]
\usepackage{times}  % DO NOT CHANGE THIS
\usepackage{helvet}  % DO NOT CHANGE THIS
\usepackage{courier}  % DO NOT CHANGE THIS
\usepackage[hyphens]{url}  % DO NOT CHANGE THIS
\usepackage{graphicx} % DO NOT CHANGE THIS
\usepackage{natbib}  % DO NOT CHANGE THIS AND DO NOT ADD ANY OPTIONS TO IT
\usepackage{caption} % DO NOT CHANGE THIS AND DO NOT ADD ANY OPTIONS TO IT
\usepackage{algorithm}
\usepackage{algorithmic}
\usepackage{times}
\usepackage{newfloat}
\usepackage{listings}
\usepackage{amsmath}
\usepackage{amssymb}
\usepackage{booktabs,makecell, multirow, tabularx} % for professional tables
\usepackage[capitalize,noabbrev]{cleveref}
\usepackage[super]{nth}
\usepackage{comment} 

\usepackage{xcolor} % Optional: To use colored text
\usepackage{soul} % Optional: To use \ul for underlining without affecting spaces
\newcommand{\BU}[1]{\textbf{\underline{#1}}}
\newcommand{\B}{\textbf}
\newcommand{\U}{\underline}
\newcommand{\quoteit}[1] {``\textit{#1}''}

\newcommand{\refappendix}[1] {#1}  % for Arxiv
\DeclareCaptionStyle{ruled}{labelfont=normalfont,labelsep=colon,strut=off} % DO NOT CHANGE THIS
\floatstyle{ruled}
\newfloat{listing}{tb}{lst}{}
\floatname{listing}{Listing}
\title{Benchmarking and Understanding Compositional Relational Reasoning of LLMs}

\author{
    Ruikang Ni\textsuperscript{\rm 1}\equalcontrib\thanks{Contribution during internship at ColorfulClouds Tech.},
    Da Xiao\textsuperscript{\rm 1}\equalcontrib\thanks{Corresponding author},
    Qingye Meng\textsuperscript{\rm 2},
    Xiangyu Li\textsuperscript{\rm 3}\footnotemark[2],
    Shihui Zheng\textsuperscript{\rm 1},
    Hongliang Liang\textsuperscript{\rm 1}
}
\affiliations{
    \textsuperscript{\rm 1}Beijing University of Posts and Telecommunications\\
    \textsuperscript{\rm 2}ColorfulClouds Technology Co., Ltd.,
    \textsuperscript{\rm 3}ICBC UBS Asset Management\\
    \{ni,xiaoda99,shihuizheng,hliang\}@bupt.edu.cn\\
    hilbertmeng@gmail.com,
    li.xiangyu@icbccs.com.cn
}

\usepackage{bibentry}
\begin{document}
\maketitle

\begin{abstract}
% AAAI creates proceedings, working notes, and technical reports directly from electronic source furnished by the authors. To ensure that all papers in the publication have a uniform appearance, authors must adhere to the following instructions.
Compositional relational reasoning (CRR) is a hallmark of human intelligence, but we lack a clear understanding of whether and how existing transformer large language models (LLMs) can solve CRR tasks.
To enable systematic exploration of the CRR capability of LLMs, we first propose a new synthetic benchmark called Generalized Associative Recall (GAR) by integrating and generalizing the essence of several tasks in mechanistic interpretability (MI) study in a unified framework. 
Evaluation shows that GAR is challenging enough for existing LLMs, revealing their fundamental deficiency in CRR.
Meanwhile, it is easy enough for systematic MI study.
Then, to understand how LLMs solve GAR tasks, we use attribution patching to discover the core circuits reused by Vicuna-33B across different tasks and a set of vital attention heads.
Intervention experiments show that the correct functioning of these heads significantly impacts task performance. 
Especially, we identify two classes of heads whose activations represent the abstract notion of true and false in GAR tasks respectively. They play a fundamental role in CRR across various models and tasks.
\end{abstract}
\begin{links}
    \link{Dataset and code}{https://github.com/Caiyun-AI/GAR}
\end{links}

\section{Introduction}
Compositional relational reasoning (CRR), the ability to reason about multiple types of relations between different entities and combine them to draw conclusions or make predictions, is a hallmark of human intelligence.
Transformer large language models (LLMs) have become the de facto backbone for foundation models due to their exceptional performance on various tasks. An important open question is whether and how existing LLMs can solve CRR tasks.

Attempting to answer this question, a number of works study the CRR capabilities of LLMs using \textit{synthetic benchmarks} \cite{weston2015towards,lake2018generalization,clark2020transformers}.
Compared with real-world benchmarks, synthetic benchmarks offer precise control over the data creation process, helping to understand the strengths and weaknesses of models on targeted tasks. Thus, they are more suitable for \textit{benchmarking} CRR, which occurs relatively rarely in real-world corpora. Besides, we also want to \textit{understand} the underlying mechanism by which the models solve CRR using \textit{mechanistic interpretability} (MI) \cite{bereska2024mechanistic} analysis.

One line of work studies compositional or multi-step reasoning \cite{dziri2024faith,press2023measuring,yang2024large,allen2023physics,allen2024physics,zhang2022unveiling,sanford2024transformers,brinkmann2024mechanistic,thomm2024limits,zhao2024exploring}.
They use synthetic tasks to reveal some fundamental limitations of LLMs on such tasks, manifested as either poor generalization when task complexities of training and test set differ \cite{dziri2024faith}, or the \textit{compositionality gap} \cite{press2023measuring}, i.e., how often models correctly answer all sub-problems but not generate the overall solution.
In these tasks, difficulty is mainly controlled by the number of reasoning steps. When the number becomes large, the over-complex input and usually poor performance of existing LLMs, especially open source ones, make in-depth MI study infeasible.
Several studies train specialized models from scratch on the proposed tasks, but the conclusions drawn from analyzing these models may not necessarily apply to existing LLMs.
Moreover, the benchmarks used by most of these works have only a single dimension of variety, namely, the number of steps, which captures only an aspect of CRR.

Another line of research uses synthetic tasks for MI study, e.g. associative recall (AR) \cite{ba2016using,fu2022hungry,olsson2022context}, knowledge recall (KR) \cite{meng2022locating,geva2023dissecting}, greater-than \cite{hanna2024does} and indirect object identification (IOI) \cite{wang2022interpretability}.
Using techniques such as path patching \cite{wang2022interpretability} and attribution patching \cite{syed2023attribution,hanna2024have}, some works discover the circuits - subgraph of the model's computation graph consisting of attention heads and MLPs - responsible for solving the tasks.
These works deepen our understanding of the general working mechanism of LLMs.
For example, the induction head mechanism found by studying AR tasks underpins the in-context learning capability of Transformer LLMs \cite{olsson2022context}.
And the IOI circuit is also generalizable to other tasks \cite{merullo2023circuit}.
While the tasks used in these works are suitable for MI study, they are usually too simple for mainstream LLMs (e.g., a model as small as GPT-2-117M can do AR and IOI tasks perfectly) to reveal any deficiency of LLMs in CRR.
They also lack variety; most current MI research is done on a \textit{single} task without any systematic study.

In summary, the lack of a CRR benchmark with both appropriate difficulty and sufficient variety in existing work hinders systematic MI studies on the CRR capabilities of LLMs.\refappendix{\footnote{An extensive review of related work is in Appendix A.}} In this paper, we make the following contributions:
\begin{itemize}
    \item We propose a new synthetic benchmark called Generalized Associative Recall (GAR) by integrating and generalizing the essence of several tasks in MI, e.g. associative recall, knowledge recall, indirect object identification (IOI), in a unified framework. GAR consists of a set of automatically generated tasks with varying forms (e.g. affirmative/negative, generation/classification) and difficulties. It is challenging enough to stress the CRR capability of mainstream LLMs, meanwhile simple enough for systematic MI study.
    \item We evaluate existing LLMs, e.g. open source Llama-2/3 7B-70B, close source GPT 3.5/4, on GAR to show that it is challenging for these LLMs despite appearing simple. Scaling helps but the compositionality gap increases, revealing fundamental deficiency of these LLMs in CRR.
    \item To understand how LLMs solve GAR tasks, we use attribution patching to discover the core circuits reused by Vicuna-33B across different tasks, and a set of vital attention heads. Intervention experiments show that the correct functioning of these heads significantly impacts task performance. Especially, we identify two classes of heads whose activations represent the abstract notion of true and false in GAR tasks respectively. Experiments show that they play fundamental roles in CRR across various models and tasks. To our knowledge, it is the first time that such heads are identified and studied in real LLMs.
\end{itemize}
% Especially, we identify two types of heads whose activations signal the presence of true and false statements respectively, thus representing the abstract notion of entailment/contextual truthfulness. Experiments show that these heads play fundamental role in CRR across various models and tasks.

\section{Generalized Associative Recall Benchmark}
\subsection{Basic Idea: From AR and KR to GAR}
Two well studied synthetic tasks in MI are AR and KR, which inspire GAR.
In AR, given a sequence of key-value ($K$-$V$) pairs as context and one of the keys as query $Q$, the model must recall the value associated with the specified key as answer $A$, e.g. \quoteit{H 1 C 4 M 7 ... C $\rightarrow$ 4}.

Compared with AR that requires the model to recall the information in context, KR requires the model to retrieve the factual knowledge stored in its parameters.
Given a subject $Q$ and a relation, the model must predict the corresponding attribute $A$, e.g. \quoteit{A dog is a kind of $\rightarrow$ animal}.

\begin{figure*}[ht]
\centering
\includegraphics[width=0.95\textwidth]{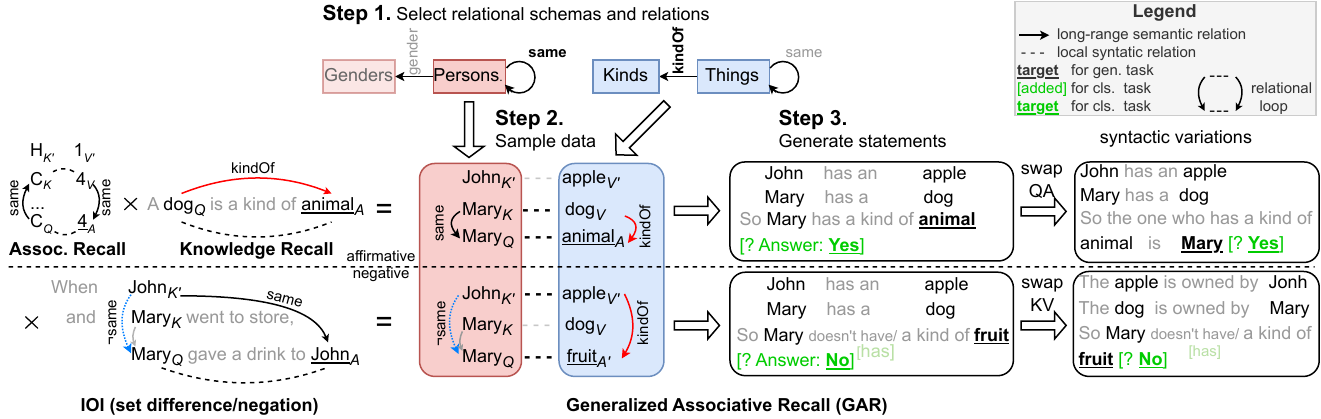}
\caption{Overall framework of Generalized Associative Recall (GAR).}
\label{fig:overall framework}
\end{figure*}

To view these two tasks from a unified perspective, we connect elements $K$, $V$, $Q$, $A$ in them with edges representing two types of relations, as shown in \cref{fig:overall framework} (left):
long-range \textit{semantic relations} (solid arrows), e.g. \texttt{same}, \texttt{kindOf}; 
local \textit{syntactic relations} (dashed lines), e.g. subject-object, adjacent positions.
We observe that for both tasks the two types of relation edges interleave and form a loop\footnote{Interleaved long-range and local relations are also characteristics in other benchmarks for reasoning, e.g. \citet{zhang2022unveiling}}, which we call a \textit{relational loop}. This is not a coincidence because, from first principles, the existence of the loop ensures predictability. 
% \citet{zhang2022unveiling,sanford2024transformers}

There are two semantic relations in AR. One (denoted as $r_{lookup}$, $C$$\rightarrow$$C$ in \cref{fig:overall framework}) is used for looking up the correct $K$-$V$ pair in context, the other ($r_{retrieve}$, 4$\rightarrow$4) is for retrieving the value $V$ and predict answer $A$. In AR, both $r_{lookup}$ and $r_{retrieve}$ are \texttt{same} relation.
To generalize AR and KR to GAR, we just replace $n_r$ of these two \texttt{same} semantic relations (black arrows) in AR with other semantic relations (red arrows) like those in KR (e.g. \texttt{kindOf}), where $0 \le n_r \le 2$,
e.g. \quoteit{John has an apple. Mary has a dog ... So Mary has a kind of $\rightarrow$ animal} ($n_r = 1$. $r_{retrieve}$ is replaced).
The number of \textit{non}-\texttt{same} semantic relations $n_r$ is a key factor for task difficulty in GAR (see \cref{fig:performance_llms_on_gar} (b)).
The relational loops remain. We argue that they are the motif of CRR. Besides guiding task design, they are also helpful for understanding the mechanism, as will be shown in \cref{fig:circuits}.

\setlength{\tabcolsep}{1mm} 
\begin{table*}[ht]
\centering
%\resizebox{.95\columnwidth}{!}{
\fontsize{9}{9}\selectfont{\begin{tabular}{llllll}
\toprule
Type        & Relational schema    & $A$ (codomain)  & $R_0$ ($\in$)   & $B$ (domain)               & $R_i$ \\
\midrule
Commonsense & GendersOfPersons     & boy, girl          & \B{isA}     & John, David, Mary, Ann, ...  & \B{same}, sameGender \\
            & KindsOfThings        & fruit, animal, ... & \B{kindOf}  & apple, banana, dog, cat, ... & \B{same}, sameKind \\
            & UsagesOfThings       & drive, write, ...  & \B{usedFor} & car, truck, pen, chalk, ... & \B{same}, sameUsage \\
            & Adjectives           &                    &             & happy, glad, fast, poor, ... & \B{synonym}, antonym \\
\midrule
Factual     & OccupationOfPersons$^*$  & Actor, author, ...   & \B{worksAsA} & Tom Hanks, Stephen King, ... & \B{same}, sameOccupation \\ 
            & CountriesOfCities    & China, France, ... & \B{inCountryOf} & Beijing, Shanghai, Paris, Lyon, ... & \B{same}, sameCountry \\
            & CountriesOfLandmarks$^*$ & China, France, ... & \B{inCountryOf} & Forbidden City, Louvre Museum, ... & \B{same}, sameCountry \\
\bottomrule
\end{tabular}
\caption{Relational schemas with form $A\xleftarrow{R_0}B\stackrel{\{R_i\}}\circlearrowright$, where $\leftarrow$ denotes a one-to-many relation. Relations used in GAR tasks are bolded. Schemas marked with $^*$ are from \citet{hernandez2023linearity}. The others are manually constructed by the authors.}
% \vskip -0.1in
\label{tab:relational schemas}}
\end{table*}

\begin{table*}[ht]
\centering
%\resizebox{.95\columnwidth}{!}{
\fontsize{9}{9}\selectfont{\begin{tabular}{lllll}
\toprule
Relational Schemas / Relations & $n_{KV}$ & \makecell[l]{Semantic\\Variation} & \makecell[l]{Syntactic\\Variation} & Example \\
\midrule
KindsOfThings/$\in$ & 0 &                   &                      & \B{Papaya} is a kind of  \BU{fruit}. \\ \hline
GendersOfPersons/=, CountriesOfCities/$\in$ & 2 &  &  swapKV & \makecell[l]{\B{Madrid} attracts \B{Michael}. \B{Bangkok} attracts \B{John}.\\ So \B{John} wants to go to a city of \BU{Thailand}} \\ \hline
OccupationOfPersons/$\in$, UsagesOfThings/$\in$ & 2 & negate  &  swapKV & \makecell[l]{The \B{biro} is \B{Frida Kahlo}'s. The \B{telephone} is \B{Meryl Streep}'s.\\ The \B{artist} does not have a thing used for \BU{communicating}} \\ \hline
GendersOfPersons/=, Adjectives/$\sim$ & 3 &  g2c &  & \makecell[l]{\B{Sarah} is \B{slow}. \B{Donna} is \B{rational}. \B{Steven} is \B{selfless}.\\Can we infer that \B{Steven} is \B{altruistic}? Answer: \U{Yes}} \\ \hline
GendersOfPersons/=, KindsOfThings/= & 3 &  g2c & swapQA & \makecell[l]{\B{Tom} has \B{car}. \B{Lisa} has \B{piano}. \B{John} has \B{sweater}.\\The one who has \B{car} is \B{Lisa}? Answer: \U{No}} \\
\bottomrule
\end{tabular}
\caption{Some GAR Tasks with examples.\refappendix{ Actual prompts used for testing LLMs are in Appendix B.}}
\label{tab:task examples}}
\end{table*}

\subsection{Workflow for Task and Example Generation}
A task of GAR and a basic form example from it with $n_{KV}$ $K$-$V$ pairs are generated in three steps:
\begin{itemize}
    \item \textbf{Step 1:} Select two relational schemas from a predefined set of relational schemas (i.e. sets enriched with some relations between elements), and for each schema select a relation to be used in the next step. This work uses seven relational schemas divided into two types (\cref{tab:relational schemas}): commonsense and factual, ensuring the variety of the tasks;
    \item \textbf{Step 2:} Sample $Q$, $K$ from the domain and codomain of $r_{lookup}$ and similarly sample $V$, $A$ from $r_{retrieve}$ to make the relational loop. Also sample $n_{KV}$$-$1 elements from the complement set of $\mathrm{domain}(r_{lookup})$ and $\mathrm{domain}(r_{retrieve})$ respectively to form $n_{KV}$$-$1 distracting $K'$-$V'$ pairs. Then shuffle the $n_{KV}$ $K$-$V$ pairs (not shown in the \cref{fig:overall framework}). 
    \item \textbf{Step 3:} Convert the data structures obtained in Step 2 into natural language statements by filling templates.
% hidden prompts added maybe in Appendix.
\end{itemize}

Besides the basic form, we apply two semantic variations and two syntactic variations to increase task variety.

\noindent \textbf{Semantic variation \textit{negate}} is closely related to IOI \cite{wang2022interpretability}, another well studied task in MI. An example is \quoteit{When John and Mary went to store, Mary gave a drink to $\rightarrow$ John`}.
Drawing the relation edges (\cref{fig:overall framework} bottom left), we observe that the essence of IOI is \textit{set complements} (essentially also the logical semantics of \textit{negation}), e.g. \{``\textit{John}'', ``\textit{Mary}''\} $-$ \{``\textit{Mary}''\} = \{``\textit{John}''\} (assuming the set \{``\textit{John}'', ``\textit{Mary}''\} is the universe. $\neg$\texttt{same} is used to denote the set complement semantic relation, which means `not same'/`except').
We incorporate this into GAR to turn an affirmative statement to its negative form by adding a $\neg$\texttt{rel} semantic relation (dotted blue arrow) to the relational loop, e.g. \quoteit{John has an apple. Mary has a dog ... So Mary doesn't have a kind of $\rightarrow$ fruit}.
The introduction of negation significantly increases task difficulty.

\noindent \textbf{Semantic variation \textit{g2c}} converts the generation task to a classification one. Instead of predicting the last missing $A$, the model must judge the truthfulness of the complete statement and predict \textit{Yes}/\textit{No} for the affirmative/negative form. Examples are given in \cref{fig:overall framework} (green text).
The generation task and its corresponding classification task share the same underlying data and relational loop. % and both require the model to predict the last token.
The former is similar to the LAMBADA benchmark \cite{paperno2016lambada}, while the latter is similar to NLI tasks, e.g. \citet{bowman2015large}, allowing us to study the difference and connection between the mechanisms underlying these two forms of tasks.

\subsubsection{Syntactic variations} \textit{swapQA} and \textit{swapKV} means swapping the order of $Q$ and $A$ or of $K$ and $V$. Examples are shown in \cref{fig:overall framework} (right).
These variations change the local syntactic relation, e.g. from subject-object/prev position to object-subject/next position. %\refappendix{\footnote{Strictly speaking, like semantic relations, syntactic relations are also \textit{directional}. But in \cref{fig:overall framework} we do not use arrows for them for simplicity}.}
We use these syntactic variations to investigate if perturbing syntactic relations like this has any impact on model performance.

\subsubsection{Tasks} The two relational schemas with selected semantic relations $r_{lookup}$ and $r_{retrieve}$, number of $K$-$V$ pairs $n_{KV}$ and the applied semantic and syntactic variations jointly define a task. We use the following format for task identifiers:

$r_{lookup}$, $r_{retrieve}$$\times$$n_{KV}$[\textit{semantic var.}](\textit{syntactic var.})

e.g. GendersOfPersons/same,KindsOfThings/kindOf$\times$3 [negate](swapKV), where \texttt{same} and \texttt{kindOf} can be abbreviated as $=$ and $\in$.
In this work, $n_{KV}$ defaults to 3. 
\cref{tab:task examples} shows some tasks with examples.
The combination of these different relational schemas and variations results in a total of 384 tasks in GAR. They have varied forms and controllable difficulty levels, while still require some shared basic capabilities of CRR, enabling systematic MI study.

\section{Evaluating LLMs on GAR}
\begin{figure*}[ht]
\centering
\includegraphics[width=0.95\textwidth]{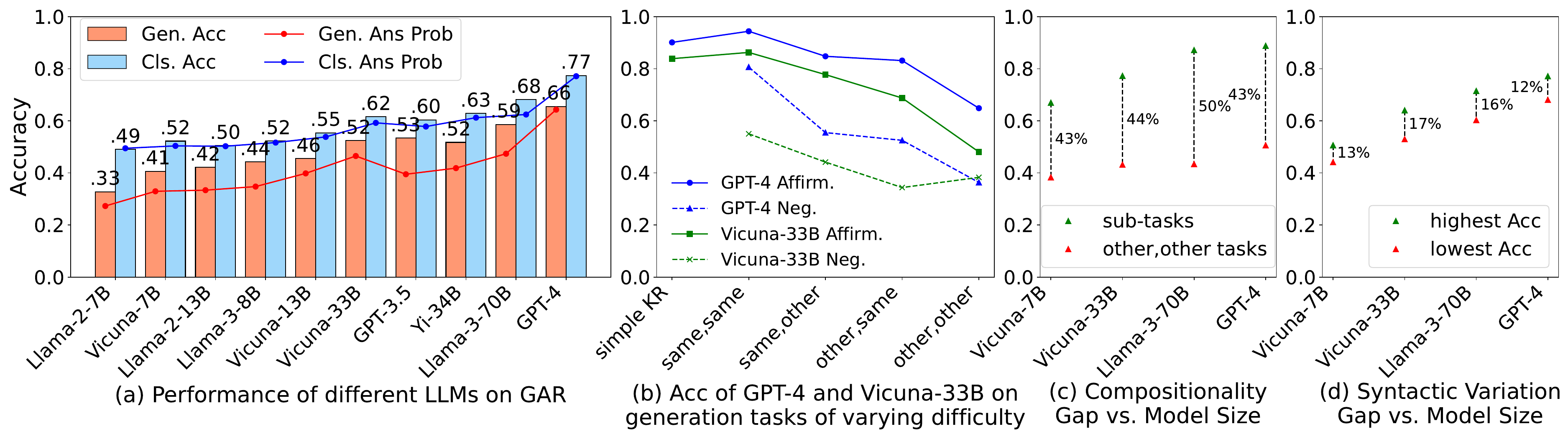}
\caption{Performance of existing LLMs on GAR.}
\label{fig:performance_llms_on_gar}
\end{figure*}

To evaluate if existing LLMs can solve GAR tasks, we test 10 models (\cref{fig:performance_llms_on_gar} (a)). The GAR dataset consists of 192 generation tasks and 192 classification tasks and a total of 4608 examples, with 8/16 examples per generation/classification task.
To obtain better performance, all examples are formatted as in-context one-shot learning.\refappendix{ The evaluation methods are described in Appendix C.}
\cref{fig:performance_llms_on_gar} (a) shows the average accuracy and predicted probability of the correct answers for both generation and classification tasks. It is evident that generation are harder than classification, and accuracy correlates positively with answer probability, both increasing with model size.
Compared with several existing multi-hop reasoning benchmarks \cite{dziri2024faith,zhang2022unveiling,sanford2024transformers}, GAR is two-hop. Though looking simple, due to the introduction of \textit{non}-\texttt{same} semantic relations and variations, GAR is still challenging for existing LLMs, even for GPT-4 with an average accuracy of only 71.5\%, far below perfect level.
%Although GPT-4 achieved the highest average accuracy at 72\%, it is still significantly below human performance levels.

\subsubsection{Task Difficulty and Compositionality Gap} GAR task difficulty can be adjusted by modifying the number of \textit{non}-\texttt{same} (other) semantic relations $n_r$ and applying the negate semantic variation. Generation task accuracies with varying difficulty are shown for GPT-4 and Vicuna-33B in \cref{fig:performance_llms_on_gar} (b)\refappendix{ and for Llama-3-70B and Vicuna-7B in Appendix C}. The accuracies on simple KR tasks with \textit{other} semantic relations (see \nth{1} example in \cref{tab:task examples}) are also shown for comparison. Task difficulty increases with $n_r$, and this trend is consistent across models of different sizes. By combining $n_r=2$ (other, other) and negate (\nth{3} example in \cref{tab:task examples}), the accuracies of the models drop below 40\%, even for GPT-4.
\cref{fig:performance_llms_on_gar} (c) shows the compositionality gaps for the above 4 models on generation tasks. We use simple KR with \textit{other} semantic relations and $n_r$=0 tasks with two \texttt{same} semantic relations (same, same), both affirmative and negative, as sub-problems since they contain enough basic knowledge and skill which can be composed to solve the hardest $n_r$=2 tasks (other, other). The compositionality gap is computed as the ratio of the average accuracy of sub-problems to that of $n_r$=2 tasks. %The gap does not decrease as the model scales up. 
The Llama series of models (including Vicuna) show an increasing compositionality gap as the model scales, revealing some fundamental deficiency of these LLMs in CRR.
In contrast, syntactic variations have much less impact on performance (\cref{fig:performance_llms_on_gar} (d)), indicating that the models are less sensitive to these perturbations.

% \begin{figure}[t]
% \centering
% \includegraphics[width=0.9\columnwidth]{acc_difficulty_gpt4_vicuna33b.pdf}
% \caption{Accuracies of GPT-4 and Vicuna-33B on \xda{generation }tasks of varying difficulty.}
% \label{fig:acc_difficulty_gpt4_vicuna33b}
% \end{figure}

\section{Discovering and Analyzing the Circuits}

\begin{figure*}[ht]
\centering
\includegraphics[width=1.0\textwidth]{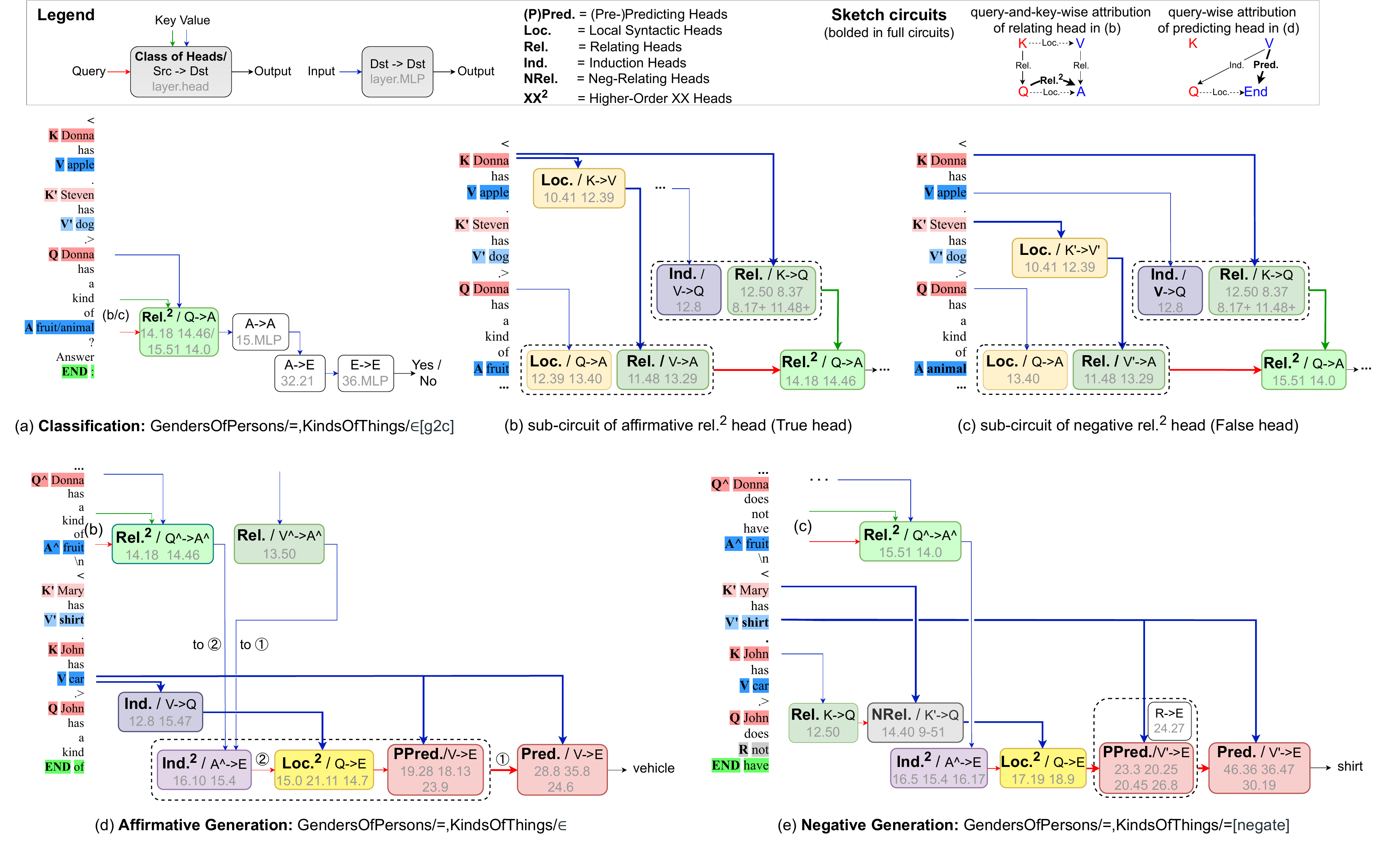}
\vskip -0.15in
\caption{Circuits of Vicuna-33B for solving GAR tasks\refappendix{. Explanations on different classes of heads are in Appendix D}}
\label{fig:circuits}
\end{figure*}

To understand the general mechanism by which LLMs solve GAR tasks, we do systematic MI study to extract reusable core circuits of Vicuna-33B by comparing individual circuits discovered for solving some tasks. We use step-wise patching similar to \citet{wang2022interpretability}, tracing from model outputs back to inputs, but replace path patching with integrated gradients-based attribution patching \cite{hanna2024have} for much faster speed. We use KL divergence as the metric to compute gradient. At each attribution step, we choose among query, key or value attribution manually, roughly following the same logic as \citet{wang2022interpretability}. Currently, we do not use any automatic circuit discovery methods because it is hard to make existing methods \cite{conmy2023towards,hanna2024have} work with our model size and circuit complexity. We leave it for future work.
We choose the Vicuna-33B model because its performance on GAR tasks is good enough for meaningful MI study while its size guarantees affordable attribution cost.
We choose 8 tasks\refappendix{ (listed in Appendix D)} that are representative and that can be solved by Vicuna-33B with high accuracy to ease attribution. Note that Vicuna-33B is larger than most models studied in MI literature with 60 layers and 52 heads. The complete circuits for these tasks are quite complex. We only show the main circuits in \cref{fig:circuits}, omitting some minor branches for clarity.

% In transformer, attention heads and multi-layer perceptrons(MLPs) play different roles in mechanistic interpretability, where the former gathers contextual information from source tokens to destination tokens, and the latter modifies destination tokens directly. 
% As shown in the legend of Figure \ref{fig:circuits}, the attention attribution of \textit{Query} and \textit{Key} in attention heads seeks to explain how one specific attention pattern forms, while the attribution of \textit{Value} and \textit{Output} shows what contextual information flows into destination tokens. However, the attribution of \textit{Input} and \textit{Output} in MLPs only reveals local transformation of destination tokens.\xda{delete this paragraph}
% In Figure \ref{fig:circuits}, we also denote critical tokens with symbols(such as Q, K, V, A), and define important heads in the legend.       

\subsection{Circuits in Classification Tasks}
% Classification tasks are easier than generation tasks for the model, resulting in relatively simple circuits. 
As shown in Figure \ref{fig:circuits} (a), the first identified contribution to \textit{Yes/No} logits comes from output of an MLP at layer 36, whose input is obtained from the hidden state of token $A$ at layer 15 through an MLP and a $A$$\rightarrow$$E$ head. A further step of attribution finds a class of very important heads, namely higher-order relating (\textit{Rel}.$^2$) heads, which includes two True heads 14.18, 14.46 and two False heads 15.51, 14.0, responsible for writing a True/False signal into the residual stream at $A$ according to the relation between $Q$ and $A$. 
% In other words, after the model sees the token A(\textit{fruit/animal}), it has already encoded truthfulness of the sentence, which later propagates to the END token. 

% Here \textit{Rel.$^2$} head denotes a higher-order relating head in comparison of common relating (\textit{Rel.}) heads, because its query and key aggregate contextual information, outputs of other attention heads, while lower-order \textit{Rel.} heads extract query and key just from intrinsic attributes of source and destination tokens(eg. \textit{apple} and \textit{fruit}).\xda{can be moved to Appendix}
% Note the order of attention heads is relative.  

Query and key attributions of the True and False higher-order relating heads are demonstrated in Figure \ref{fig:circuits} (b) and \ref{fig:circuits} (c) separately. For True heads, their query $A$ gathers the information of $Q$ by local heads, and of $K$-$V$ pair by composition of the relating and local heads, while its key $Q$ gathers the information of $K$ and $V$ by the relating and induction heads respectively. Therefore, the information of query side $K$-$V$ pair and the key side $K$-$V$ pair can be compared and matched to affirm that the statement is true. False heads (Figure \ref{fig:circuits} (c)) share similar circuit as True heads except that 1) the relating heads move the distractive $K'$-$V'$ to $A$; 2) the False heads are activated at $A$ by comparison of query-wise $K'$-$V'$ and key-wise $K$-$V$ pairs, which don't match. In both affirmative and negative circuits, the local $Q$$\rightarrow$$A$ heads contribute to the formation of $Q$$\rightarrow$$A$ attention pattern of the higher-order relating heads.
% Local Syntactic(\textit{Loc.}) Heads use syntactic relations to attend to tokens, such as \textit{Donna}$\rightarrow$\textit{apple} in ``\textit{Donna has apple}''. 
% To summarize, the critical forward circuit can be divided into three parts: 1) token A(\textit{fruit/animal}) causes \textit{KindsofThings Rel.} heads 11.48, 13.29 attending token \textit{V-apple/V'-dog}; 2) token V/V' activates True/False \textit{Rel.$^2$} heads to inject truthfulness signal; 3) the signal propagates to the END(\textit{:}) token, adjusting \textit{Yes/No} logits.

\subsection{Circuits in Generation Tasks}
Circuits in the affirmative and negative generation tasks are depicted in Figure \ref{fig:circuits} (d) and (e) separately. For the affirmative one, we first identify two classes of heads, predicting and pre-predicting heads. The former is responsible for correctly attending to $V$ (\quoteit{car}) and retrieving its \texttt{kindOf} attribute (\quoteit{vehicle}), and its query attribution reveals that the formation of its attention pattern relies on the pre-predicting heads. Further query attribution of these (pre-)predicting heads finds the higher-order local (\textit{Loc}.$^2$, $Q$$\rightarrow$$E$) heads, whose value attribution in turn finds the induction heads ($V$$\rightarrow$$Q$), through which the information of $V$ flows through $Q$ into $End$, helping form the attention $V$$\rightarrow$$E$ of the predicting heads.
Query attribution of the \textit{Loc}.$^2$ heads discovers another class of higher-order Induction (\textit{Ind}.$^2$) heads, which transmit in-context signals from answer $A\string^$ in the previous one-shot demonstration to the current token, promoting the activation of the \textit{Loc}.$^2$ heads ($Q$$\rightarrow$$E$) by the higher-order True heads ($Q\string^$$\rightarrow$$A\string^$) activated on the previous demonstration (path \textcircled{2}), which are exactly the same class of heads we find in the classification circuits (\cref{fig:circuits} (b)).
Similarly, the relating head ($V\string^$$\rightarrow$$A\string^$) activated on the demonstration promotes the activation of the predicting heads $V$$\rightarrow$$E$ (path \textcircled{1}), through the same set of \textit{Ind}.$^2$ heads. 
So the higher-order induction heads play a fundamental role in bridging different classes of attention heads across context\refappendix{, which is also validated by circuits in other generation tasks shown in Appendix D}.

% Similarly, query attribution of \textit{PPred.} heads unveils \textit{KindsOfThings Rel.} head(V\^{}$\rightarrow$A\^{}) via the same \textit{Ind.$^2$} heads. 

The predicting heads identified in the negative task are different from those in the affirmative task because $r_{retrieve}$ changes from \texttt{KindsOfThings} to \texttt{same}.
With similar attribution we can find different pre-predicting, higher-order local and higher-order relating heads consecutively. Value attribution of the higher-order local heads shows that a relating head ($K$$\rightarrow$$Q$) promotes negative-relating heads ($K'$$\rightarrow$$Q$) which transmit the information of $K'$ to $Q$ then to $E$, helping the (pre-)predicting heads attend from $E$ to $V'$, which also has the information of $K'$.
An $R$$\rightarrow$$E$ head attending to \quoteit{not} also has some contribution to the attention formation of the predicting heads.
The key difference between the negative and affirmative circuits is that different higher-order relating heads activated on the demonstration activate different higher-order local heads, though via the same higher-order induction heads. 

\subsection{Core and Sketch Circuits}

% both $A$ and $Q$ aggregate ($K$, $V$) pair by \textit{Loc.} heads and \textit{Rel.} heads, forming \textit{Rel.$^2$} and completing the higher-order relational loop. 
% Therefore, conjunction of circuits can help construct relational loops and capture hidden relation in the context, on top of which higher-order heads and relational loops form, representing more complex contextual relations. 

% The query side forms a first-order loop by induction heads, with which \textit{Rel.$^2$} heads compare the key side pair, forming the second-order loop.

% \subsection{Core Circuits} 
Putting it all together, we can extract the core circuits reused across various tasks.
\begin{itemize}
    \item \textbf{Truthfulness Sub-Circuit:} The circuit of higher-order relating heads (Figure \ref{fig:circuits} (b) and (c)) shared between classification and generation tasks detects the higher-order relation between $A$ and $Q$ to judge truthfulness. 
    \item \textbf{ICL Sub-Circuit:} Across various generation tasks, different (higher-order) relating heads activate different higher-order local heads or predicting heads through the same higher-order induction heads, which plays a fundamental role in in-context learning. 
    \item \textbf{Overall Circuit:} The truthfulness sub-circuit can be combined either with downstream MLPs to complete classification tasks (\cref{fig:circuits} (a)), or with downstream attention heads via the ICL sub-circuit to solve generation tasks (\cref{fig:circuits} (d), (e)).
    % which indicates that LLMs tend to unify and simplify circuits in different tasks to integrate a backbone circuit, combining attention heads and MLPs depending on the context.
\end{itemize}

To understand these circuits from the \textit{relational loop} perspective, 
the first sketch in Figure \ref{fig:circuits} (legend) shows query-wise (from $A$) and key-wise (from $Q$) attribution of the higher-order relating heads in Figure \ref{fig:circuits} (b), where two links between $Q$ and $A$ appear: one is formed directly by the local heads; the other is formed indirectly by composition of query-wise $K$$\rightarrow$$V$$\rightarrow$$A$ and key-wise $K$$\rightarrow$$Q$. The two links together form a relational loop, which are \textit{detected} by the higher-order True heads. The second sketch shows a high-level sketch of the circuit in Figure \ref{fig:circuits} (d). The critical attention pattern $V$$\rightarrow$$E$ of the predicting heads is formed by composition of $V$$\rightarrow$$Q$ and $Q$$\rightarrow$$E$, which \textit{completes} the relational loop required for solving the generation task. 

\section{Validating Attention Heads}
In this section, we use intervention and other methods to validate the vital roles of several types of heads identified in the previous section. We first explain a few concepts:
\begin{itemize}
\item \textbf{Head Activation}: If at any attending position (a row in the attention weight matrix) a head assigns the largest attention weight to any token other than the first token \verb|<|$s$\verb|>|, this head is considered to be \textit{activated}.\footnote{We observed that for Llama/Vicuna models the first token plays the role of ``sink token'' for null attention \cite{xiao2023efficient}.} The largest such weight across the whole attention matrix is defined as its \textit{activation value}.
\item \textbf{True/False Heads}: Higher-order relating heads which activate on true/false statements for truthfulness judgement (\cref{fig:circuits} (b) / (c)).
% \item \textbf{False Heads}: Contrary to True heads, these heads also accurately distinguish between "Yes" and "No" answers but assign more attention when the answer is "No."\xda{merged with True heads?}
\item \textbf{Strong Intervention}: This intervention forces the attention weights of a head to fully comply with the expected attention pattern, e.g. for attention pattern $V$$\rightarrow$$E$, the attention weight of token $End$ attending to token $V$ is set to 1 while the weights to the other tokens are set to 0.
% \item \textbf{Weak Intervention}: This intervention finds the maximum attention \xdr{score}{weight} corresponding to the pattern within the entire model, replaces the target head's attention score at that pattern with this score.\xda{not clear enough} Unlike strong intervention, which directly imposes an attention pattern on the model, weak intervention utilizes the attention patterns already learned by the model, thus avoiding the introduction of additional information \xdr{and errors}{from outside the model}.
\item \textbf{Weak Intervention}: This intervention replaces the attention weights of a head at the attending position (e.g. $End$) with weights found from all other heads in the model that complies best with the attention pattern (e.g. $V$$\rightarrow$$E$).\refappendix{\footnote{See illustrations of strong/weak interventions in Appendix E.}} Unlike strong intervention which directly imposes an attention pattern (essentially the same as the intervention method used in \citet{merullo2023circuit}), weak intervention utilizes the attention weights already formed by the model itself, though from other heads, thus avoiding the introduction of additional information from outside the model.
\end{itemize}

\subsection{Validating non-True/False Heads in Vicuna-33B}
%In this section, we explore the fundamental and universal attention patterns by intervening in the lower-level heads' attention. The activation of True/False relational loops might be triggered by the output information from these heads. Understanding the role of these lower-level heads is crucial for a comprehensive understanding of the model's working mechanism.
\begin{figure*}[ht]
\centering
\includegraphics[width=1.0\textwidth]{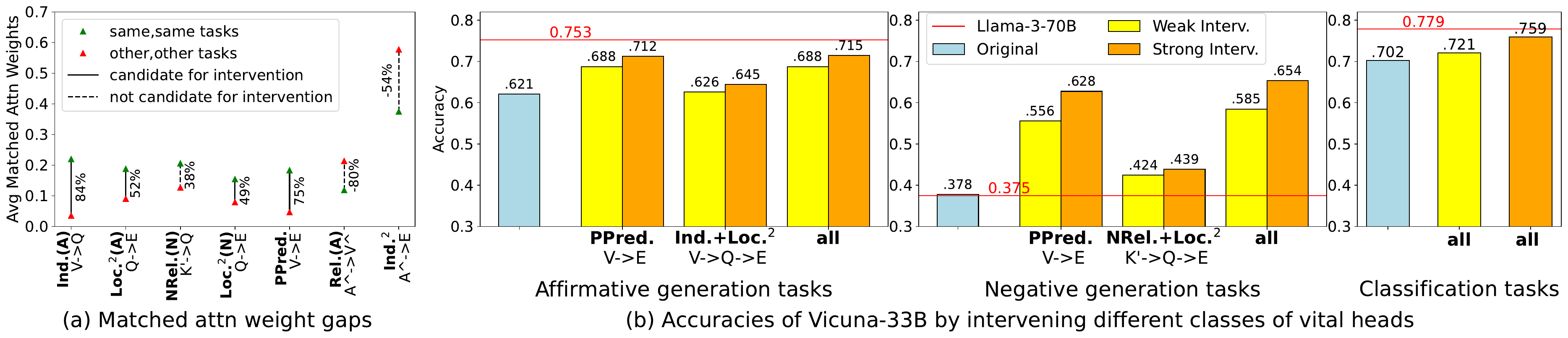}
\caption{Analysis and intervention results of some vital heads in Vicuna-33B. In Figure (a), (A) denotes affirmative generation tasks and (N) denotes negative generation tasks.}
\label{fig:intervention_non_true_false_heads}
\end{figure*}

\begin{figure}[ht]
\centering
\includegraphics[width=1.0\columnwidth]{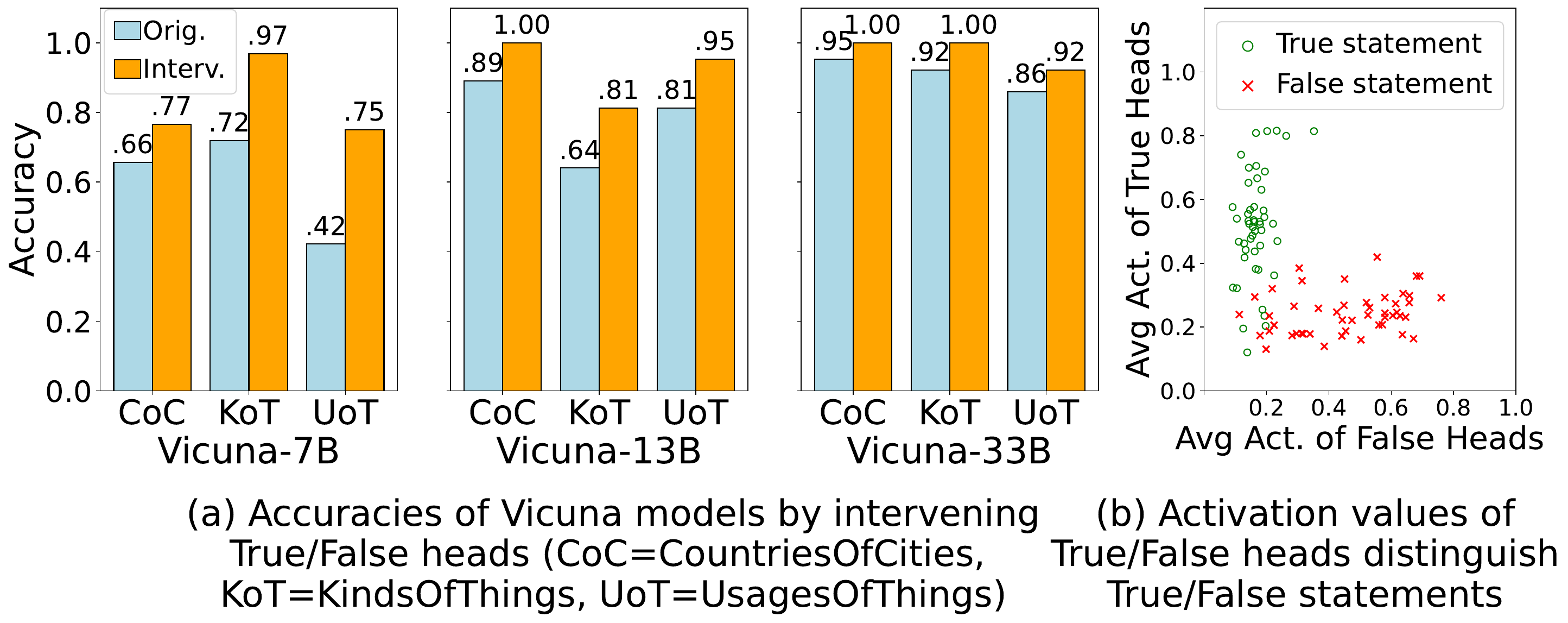}
\caption{Analysis and intervention of True/False heads.}
\label{fig:intervention_true_false_heads}
\end{figure}

% \noindent \B{Affirmative generation tasks:} \\
% - Pre-Predicting Heads: ($V$$\rightarrow$$E$) 19.28, 18.13, 23.9 \\
% - Induction Head: ($V$$\rightarrow$$Q$) 12.8, 15.47, 16.33 \\
% \B{Negative generation tasks:} \\
% - Pre-Predicting Heads: ($V$'$\rightarrow$$E$) 20.25, 20.45, 23.3, 26.8 \\
% - \mqyr{Different Token}{Negative Relating} Heads: ($K$'$\rightarrow$$Q$) 14, 40 \\
% \B{Classification tasks:} \\
% - Relating Heads: ($V$$\rightarrow$$A$) 11.48, 13.29, ($K$$\rightarrow$$Q$) 12.50, 8.37 \\
% - Induction Heads: ($V$$\rightarrow$$Q$) 12.8
% These heads exhibited highly generalized and robust attention patterns in our attribution analysis, demonstrating significant attention values across multiple tasks and making significant contributions to the model's output. 

We apply weak and strong interventions to several classes of attention heads in \cref{fig:circuits} to validate their importance.
As a motivation for intervention, we first inspect their average attention weights in generation tasks that match the desired attention pattern, e.g. the average weight of $End$ attending to $V$ for pattern $V$$\rightarrow$$E$, which measures how well the heads function as expected.
As shown in \cref{fig:intervention_non_true_false_heads} (a), similar to the compositionality gap, the matched attention weights of some classes of heads also exhibit a clear gap between the easiest same,same tasks and the hardest other,other tasks (recall discussion on task difficulty and compositionality gap in \cref{fig:performance_llms_on_gar} (c) for definition of these tasks), partially explaining the performance drop on harder tasks.
In contrast, higher-order induction heads are more robust, maintaining high matched attention weights across tasks.

% how to intervene 
For generation tasks, we intervene on the class of heads with positive matched attention weight gaps in \cref{fig:intervention_non_true_false_heads} (a). We jointly intervene on higher-order local heads with induction heads (Ind.+Loc.$^2$, for affirmative tasks) or with negative-relating heads (NRel.+Loc.$^2$, for negative tasks), because they V-compose to form critical paths in the circuits (\cref{fig:circuits} (d,e)). % We also intervention all of them simultaneously.
For classification tasks, we jointly intervene on all relating heads and induction heads, because they work in parallel to form the attention of True/False heads (\cref{fig:circuits} (b,c)).
% All classes of heads we intervene are universal, i.e. remain relatively consistent across tasks.
We don't intervene on 1) predicting heads, which vary between different $r_{retrieve}$ for generation tasks and are thus not universal, and 2) relating heads in affirmative generation tasks and higher-order induction heads, which exhibit negative matched attention weight gaps.

% Results
\cref{fig:intervention_non_true_false_heads} (b) shows that strong intervention jointly on all heads increase the accuracy of Vicuna-33B significantly, approaching (for affirmative generation and classification tasks) or even surpassing (for negative generation tasks) the performance of Llama-3-70B. % The interventions had a more significant impact on the task variants which are more difficult.
For generation tasks, intervention on pre-predicting heads is most effective, which is intuitive because they are closest to and thus have the most direct impact on the final output of the model.
Intervening on the other heads (Ind.+Loc.$^2$ in affirmative tasks and NRel+Loc.$^2$ in negative tasks) has smaller effect, perhaps because they are too far away from the final output to have strong impact.
To conclude, the effectiveness of strong intervention shows that the correct functioning of these heads, especially correct formation of the required attention patterns, significantly impact task performance.
Weak intervention is also effective, albeit weaker, indicating that the model has the intrinsic ability to attend correctly, but with other heads that cannot compose the functional circuits, showing potential for improvement.

\subsection{Validating True/False Heads across Models}
% Building on the experiments from Section X.1, we further validate the universality and effectiveness of the \xdr{attention }{}True/False heads through intervention\xdd{ and knockout operations}. Given the critical importance of these heads in classification tasks, we conducted experiments specifically focused on them, using three \xdd{fundamental and representative }tasks \xda{list task ids }to demonstrate their effectiveness. 
After validating the upstream heads of the True/False heads in Vicuna-33B (\cref{fig:intervention_non_true_false_heads} (b) rightmost), we further validate the effectiveness and universality of the True/False heads themselves by intervention on different-sized vicuna models. We conduct experiments on three representative tasks\refappendix{ used for circuit discovery in Appendix D}: GendersOfPersons/=, $r_{retrieve}/\in$, where $r_{retrieve} \in$ \{CountriesOfCities (CoC), KindsOfThings (KoT), UsagesOfThings (UoT)\}.
Using the attribution method described previously, we identify the True/False heads in different-sized Vicuna models\refappendix{ (listed in Appendix E)} for intervention.
Specifically, when the answer is \textit{Yes}/\textit{No}, we apply intervention on the True/False heads while knocking out the False/True heads.
As shown in \cref{fig:intervention_true_false_heads} (a), after intervention, the average accuracy of Vicuna-7B/13B/33B is increased by 17\%/\/14\%/6\%, indicating that the True/False heads are universal and exhibit consistent effects across different model sizes.

To understand why Vicuna-33B already has high classification accuracy before intervention, we randomly sample 96 examples from the classification tasks and plot each example as a dot using the average activation values of the True heads and False heads as coordinates in \cref{fig:intervention_true_false_heads} (b).
It can be seen that the activation values can effectively distinguish true and false statements, indicating that these True/False heads definitely represent the abstract notion of true and false in these tasks.
Combining \cref{fig:intervention_true_false_heads} (a) and (b), it is evident that the activation status of True/False heads encodes the truthfulness of statements and that the models indeed use them to judge truthfulness in GAR tasks.

\subsection{The efficacy of True/False Heads in Other Datasets}
To investigate if the True/False heads also play an important role on other datasets besides GAR, we test them on two datasets that require to judge if a statement is true (entailment) or false (contradiction): Stanford Natural Language Inference (SNLI, \citet{bowman2015large} and Geometry of Truth (GoT, \citet{marks2023geometry}.
We forward the examples through Vicuna-33B and extract the activation values of four True/False heads (14.18, 14.46, 15.51, 14.0) as features to train simple MLP classifiers to predict true or false.
\refappendix{The experimental detail is in Appendix E.}

\cref{tab:acc_mlp_true_false} shows the results along with one-shot accuracies of several other different-sized models on these two datasets for comparison. The MLP classifiers' accuracies approach (SNLI) or surpass (GoT) Vicuna-7B, indicating that the activations of these heads encode information for truthfulness classification. The activation patterns of these True/False heads identified in GAR are robust across other datasets.

% \xda{add results for randomly sampled heads}
\setlength{\tabcolsep}{1mm} 
\begin{table}[ht]
\centering
\fontsize{9}{9}\selectfont{\begin{tabular}{lll}
\toprule
Model  & SNLI Acc(\%)  & GoT Acc(\%)  \\
\midrule
Random Guess         & 50    & 50     \\
GPT-2-Medium (345M)  & 51.43 & 51.04  \\
GPT-2-XL (1.5B)      & 53.90 & 50.35  \\
% \B{True/False Head Act + LR} & \B{83.53\%($\pm$1.5\%)} & \B{70.43\%($\pm$10.2\%)}\\
\B{True/False Head Act + MLP} & \B{87.07$\pm$0.2} & \B{85.63$\pm$1.8}\\
Vicuna-7B            & 91.00 & 84.73  \\
Vicuna-33B           & 96.80 & 89.69  \\
\bottomrule
\end{tabular}
\caption{Accuracy of different models and MLP classifiers with activation values of True/False heads of Vicuna-33B as features on SNLI and on 4 subsets of GoT.}
\label{tab:acc_mlp_true_false}}
\end{table}

\section{Conclusion}
We propose the Generalized Associative Recall (GAR) benchmark %by integrating and generalizing the essence of several tasks in MI, e.g. associative recall, knowledge recall, indirect object identification (IOI), in a unified framework.
that is challenging enough to stress the CRR capability of mainstream LLMs, meanwhile simple enough for systematic MI study.
We evaluate existing LLMs on GAR to show that the compositionality gap increases despite scaling, revealing fundamental deficiency of these LLMs in CRR.
We discover the core circuits reused by Vicuna-33B across different GAR tasks and a set of attention heads important for task performance, especially the True/False heads, which can represent true/false statements in GAR tasks and play fundamental roles in CRR across various models and tasks.

\section{Acknowledgements}
The work was partially supported by National Natural Science Foundation of China (NSFC) (Grant No. 62425105, 62350001, 62206019), Fundamental Research Funds for the Central Universities (Grant No. 530424001) and Taiyuan City ``Double hundred Research action'' 2024TYJB0127.
\bibliography{aaai25}

\newpage
\appendix

\section{A. Related Work}
% \textbf{Benchmarks on Compositional and Relational Reasoning.}
% As compositional and relational reasoning (CRR) is a critical ability for real-world applications of LLMs, a number of works focus on benchmarking \cite{dziri2024faith,press2023measuring,thomm2024limits} multi-hop or more complex reasoning of LLMs and investigating limitations \cite{dziri2024faith,thomm2024limits}, improvement \cite{press2023measuring, biran2024hopping} or understanding the mechanism \cite{brinkmann2024mechanistic,allen2023physics,allen2024physics} of transformer in tackling multi-hop reasoning tasks. 

\subsection{Research on Compositional and Relational Reasoning of Transformer LLMs}
As compositional and relational reasoning (CRR) is a critical ability for real-world applications of LLMs, a number of works focus on benchmarking \cite{dziri2024faith,press2023measuring,thomm2024limits}, improving \cite{press2023measuring, biran2024hopping} or understanding the mechanism of \cite{brinkmann2024mechanistic,allen2023physics,allen2024physics} transformer LLMs in tackling compositional or multi-step reasoning tasks. 

% With regards to benchmarking, 
\citet{dziri2024faith} investigated the limits of closed source LLMs (eg. GPT4) across three representative compositional tasks (multi-digit multiplication, logic grid puzzles, and a classic dynamic programming problem), and demonstrated that as problem size increases, accuracy of LLMs decreases near to zero. \citet{press2023measuring} evaluated two-hop reasoning and proposed compositionality gap by measuring how often models can correctly tackle all sub-problems but fail to generate the overall solution. 
% Although accuracy of single-step and multi-step reasoning improves as we scale model size, the compositionality gap still remains.
Experiments by \citet{thomm2024limits} showed that compositional learning in Transformer language models is highly sample inefficient. 
In accordance to these results, we also find that LLMs struggle on the difficult part of GAR tasks and even GPT-4 is far from perfect performance.

% \xda{remove paragraph} To improve multi-hop reasoning, some methods are proposed, including eliciting prompting (self-ask) \citep{press2023measuring}, back-patching \citep{biran2024hopping}, which use more tokens to alleviates the problem that the order of hops in multi-hop reasoning does not match corresponding attention heads or MLPs along transformer layers. In our work, we can improve accuracy by intervening on critical heads in the circuits.

\citet{zhao2024exploring} proposed Multilingual Compositional Relation (MCR) benchmark, which is very relevant to GAR. While both benchmarks are proposed to study CRR, they focus on different aspects of CRR and are complementary:
1) They compose relations in different ways. GAR uses $r_{lookup}$ to locate V by attention among $n_{KV}$ distractive candidates V's, then uses $r_{retrieve}$ to retrieve attribute from V. In MCR both relatoins R and S act like $r_{retrieve}$ and compose like "son x son = grandson". There is no attention lookup from distractors.
2) MCR has multilingual and definition/reasoning variations and controls difficulty by the number of relations. GAR has negate and g2c semantic variations and swapQA/KV syntactic variations and controls difficulty by the number of non-same relations and adding negate.
3) While GAR is generated programmatically from template, MCR is human curated from natural text and more diverse.

When analyzing LLMs trained on natural languages, \citet{yang2024large,biran2024hopping} found LLMs perform two-hop reasoning latently for the prompts of certain relation types. Instead of GPT, \citet{zhang2022unveiling} studied how BERT and ALBERT learn to reason on LEGO benchmark that encapsulates the problem of following a chain of reasoning. \citet{allen2023physics,allen2024physics} investigated four knowledge manipulation tasks and uncovered the hidden reasoning mechanism by which language models solve grade school math word problems.  
From a theoretical perspective, \citet{sanford2024transformers} proved that a constant number of self-attention layers can efficiently simulate—and be simulated by—a constant number of communication rounds of Massively Parallel Computation. Meanwhile, they validated experimentally that k-hop induction heads task can be solved by transformer of depth $\mathcal{O}(\log k)$. Similarly, in mechanistic analysis of transformer trained on a synthetic reasoning task (path-finding), \citet{brinkmann2024mechanistic} also found some parallelization mechanism. 

Although experiments on well-controlled dataset appear scientific and rigorous, some conclusions drawn from analyzing models trained from scratch on these synthetic tasks \cite{allen2023physics,brinkmann2024mechanistic} may not apply to existing LLMs well, due to the difference between curated synthetic data and noisy natural texts and between model sizes. We explore general working mechanisms of existing LLMs (eg. Vicuna-33B) on the GAR synthetic tasks, which is of more practical importance. In addition, existing works lack in-depth MI analysis of LLMs on compositional and relational reasoning. Some of these tasks (eg. logic grid puzzles) are too complicated for open-source models to make MI study. Some of them are less difficult: the two-hop reasoning in works \cite{yang2024large,biran2024hopping} can be treated as two consecutive knowledge recalls, which is simpler than our GAR tasks.

\subsection{Mechanistic Interpretability on Specific Tasks}
MI seeks to reverse-engineer neural networks by localizing circuits \cite{olah2020zoom}, computational subgraphs of neural networks for solving specific tasks. Previous works studied MI on a variety of tasks, such as associative recall (AR) \cite{ba2016using,fu2022hungry,olsson2022context}, knowledge recall (KR) \cite{meng2022locating,geva2023dissecting}, gender-bias \cite{chintam2023identifying}, capital-country \cite{merullo2024language}, subject-verb agreement, hypernymy \cite{hanna2024have}, greater-than \cite{hanna2024does}, indirect object identification (IOI) \cite{wang2022interpretability} and colored objects \cite{merullo2023circuit}. 

In comparison of above works, we do systematic MI study of larger models on a set of different but related tasks in both zero-shot and few-shot settings. In terms of tasks, GAR is a more general framework, subsuming AR, KR and IOI as special cases. In term of circuits discovered, ours are more complex, because of increased difficulty by few-shot generation and composition of AR and KR. Several classes of heads in our circuits are analogous to those found in other works. For example, predicting heads are a general form of name mover heads in \citet{wang2022interpretability}. Additionally, we discover new types of heads, higher-order ones, especially True/False heads. 

Induction heads \cite{elhage2021mathematical}, are a very powerful mechanism for achieving in-context learning. Recent studies \cite{todd2023function,hendel2023context,liu2023context} show that
LLMs can extract function/task vectors from few-shot examples. Using causal mediation analysis on ICL tasks, \citet{todd2023function} identified a small number attention heads transporting a compact representation of the previous demonstrations. In contrast to previous works, we identify the higher-order induction heads and extract the general ICL circuit across tasks, in which relating heads activate predicting heads through the higher-order induction heads. 

% \xda{Remove paragraph }\textbf{Attribution Methods.}
% Main attribution methods used in MI include probing, path patching \cite{wang2022interpretability} and attribution patching \cite{syed2023attribution,hanna2024have}, Automatic Circuit DisCovery (ACDC) \cite{conmy2023towards}, and edge attribution patching with integrated gradients (EAP-IG) \cite{hanna2024have}. In contrast to these methods, we use step-wise patching similar to \citet{wang2022interpretability}, tracing from model outputs back to inputs, but replace path patching with integrated gradients-based attribution patching \cite{hanna2024have} for much faster speed. At each attribution step, we choose among query, key or value attribution manually, roughly following the same logic as \citet{wang2022interpretability}. Currently, we do not use any automatic circuit discovery methods because it is hard to make existing methods \cite{conmy2023towards,hanna2024have} work with our model size and circuit complexity. 
% Attribution of large models (eg. Vicuna-33B) on complex tasks is challenging for existing automatic attribution methods,
\setlength{\tabcolsep}{1mm} 
\begin{table*}[h!]
\centering
\fontsize{9}{9}\selectfont{\begin{tabular}{lllll}
\toprule
Relational Schemas / Relations & $n_{KV}$ & \makecell{Semantic\\variation} & \makecell{Syntactic\\variation} & Example \\
\midrule
\makecell[l]{GendersOfPersons/=,\\CountriesOfCities/$\in$} & 2 & & swapKV & \makecell[l]{\textit{Countries of cities include Spain, Thailand, Italy and Russia.}\\ $<$Petersburg attracts Sharon. Milan attracts Barbara.$>$.\\ So Barbara wants to go to a city of Italy\\ $<$Madrid attracts Michael. Bangkok attracts John.$>$.\\ So John wants to go to a city of \U{Thailand}} \\ \hline
\makecell[l]{GendersOfPersons/=,\\KindsOfThings/=} & 3 & g2c & swapQA & \makecell[l]{\textit{Premise:} $<$Sandra has pants. Kevin has cannon. Alice has peach.$>$.\\ \textit{Answer with Yes or No.} Can it be inferred \textit{from the premise} that\\ the one who has peach is Alice? Answer: Yes\\ \textit{Premise:} $<$Tom has car. Lisa has piano. John has sweater.$>$.\\\textit{Answer with Yes or No.} Can it be inferred \textit{from the premise} that\\ the one who has car is Lisa? Answer: \U{No}} \\
\bottomrule
\end{tabular}
\caption{Actual prompts used for the \nth{2} and \nth{5} examples in \cref{tab:task examples}. Main differences are italicized.}
\label{tab:prompt_details}}
\end{table*}

\section{B. Details of Prompts Used in GAR Tasks}
\cref{tab:prompt_details} shows by examples the actual prompts used in GAR tasks with the following differences compared with those in \cref{fig:overall framework} and \cref{tab:task examples}, which empirically improve performance of LLMs:
\begin{itemize}
    \item prepending a list of randomly shuffled candidate answers when $r_{retrieve} \neq$ \texttt{same} e.g. \quoteit{Countries of cities include Spain, Thailand, ...};
    \item bracketing the $K$-$V$ pairs with ``\verb|<| ... \verb|>|'';
    \item labeling and referring to the $K$-$V$ pairs as the \quoteit{premise} and adding a prompt \quoteit{Answer with Yes or No.} in classification tasks;
    \item adding one-shot demonstration.
\end{itemize}

\section{C. Details of Evaluating LLMs on GAR}
The Llama-2-7B/13B, Llama-3-8B/70B, and Yi-34B models are accessed via together.ai completion API\footnote{https://api.together.xyz/v1/completions}. The input text includes the target to be predicted, and the logprob of the token at the target position is used to calculate its probability. If this probability exceeds the average probability of the alternative answers (33.3\% for generation tasks and 50\% for classification tasks), the model is considered to have answered correctly. The gpt-3.5-turbo-0125 and gpt-4-turbo-2024-04-09 models are accessed via OpenAI chat API, with input text that does not include the target. An additional prompt is prepended to constrain the model’s output: ``Output a word or phrase to complete the last sentence according to the given examples.'' Since the output from chat models is usually not a complete word, we consider the first token in the output that matches the prefix of the correct answer as the correct response.  
The Vicuna-7B/13B/33B-v1.3 models, which are corresponding Llama models finetuned on conversation data, are deployed and tested locally using Huggingface Transformers library. The input text does not include the target, and the output logits from the last position are used to compute the probability of the target.

\begin{figure}[ht]
\centering
\includegraphics[width=0.9\columnwidth]{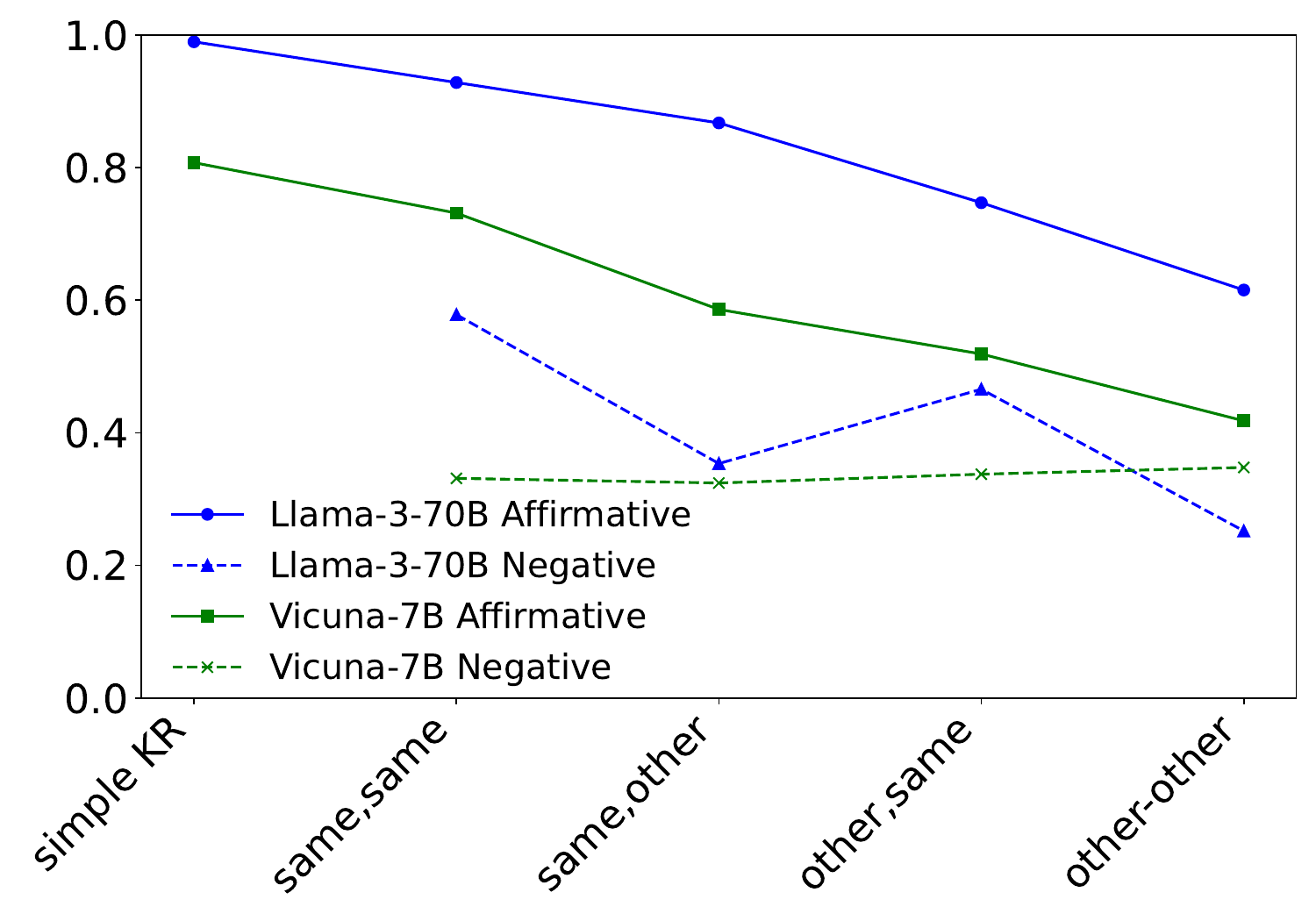}
\caption{Accuracies of Llama-3-70B and Vicuna-7B on tasks of varying difficulty.}
\label{fig:Llama_vicuna_acc_on_vary_difficulty}
\end{figure}

\subsection{Error analysis}
We manually analyze 100 randomly sampled error cases of generation tasks from GPT-4 and Vicuna-33B. The errors fall in 3 classes (For example: \quoteit{John has apple. Mary has dog. Tom has lemon. So the boys don't have a kind of $\rightarrow$ animal}): 
\begin{enumerate}
    \item synonym of correct answer (can be treated as correct), e.g. "pet";
    \item answer by applying a non-same $r_{retrieve}$ to a wrong candidate V', e.g. "fruit";
    \item answer not relevant to any candidates, maybe due to lack of knowledge or hallucination, e.g. "clothes".
\end{enumerate}
As shown in \cref{tab:error_analysis}, for both models Class 2 errors are dominant, indicating that the models fail to correctly compose $r_{lookup}$ and $r_{retrieve}$ and have fundamental deficiency on the tasks.
\setlength{\tabcolsep}{1mm} 
\begin{table}[H]
\centering
\fontsize{9}{9}\selectfont{\begin{tabular}{llll}
\toprule
Model      & Class 1 & Class 2 & Class 3   \\
\midrule
GPT-4      & 21      & 79      & 0  \\
Vicuna-33B & 6       & 88      & 6  \\
\bottomrule
\end{tabular}
\caption{Number of 3 classes of errors for two models.}
\label{tab:error_analysis}}
\end{table}

\section{D. More on Discovering and Analyzing the Circuits}
\subsection{Explanations on Types of Heads in Circuits}
In classification and generation circuits, we find a few classes of attention heads which are important for solving the tasks. We can further relate these heads to the semantic or syntactic relations in the GAR framework to better understand their roles.

\begin{itemize}
    \item \textbf{Predicting Heads} output the answer by copying or retrieving the attribute of token $V$ in accordance with semantic relation $r_{retrieve}$. For example, in \cref{fig:circuits} (d), the predicting heads output \quoteit{vehicle} from \quoteit{car} with KindsOfThings/\texttt{kindsOf} semantic relation.
    \item \textbf{Pre-Predicting Heads} promote the attention formation of predicting heads by attending the same token (eg. \quoteit{car} in the above example) as predicting heads.
    \item \textbf{Local Syntactic Heads} attend between syntactically related tokens, e.g., attending from \quoteit{apple} to \quoteit{Donna} in the context of \quoteit{Donna has apple}, based on the subject-object syntactic relation. 
    \item \textbf{Relating Heads} form attention patterns according to semantic relations (either $r_{lookup}$ or $r_{retrieve}$). For instance, \quoteit{fruit} attends to \quoteit{apple} with relation \texttt{kindsOf}, and \quoteit{Donna} to \quoteit{Donna} with relation \texttt{same}. They can be seen as a general form of duplicate token heads in IOI \cite{wang2022interpretability},  which signal the occurrence of token duplication. Relating heads in GAR signal the occurrence of the semantic relation.
    \item \textbf{Induction Heads} 
    % predict the same token which relates (with some relation $r_{1}$) to the token that has some relation $r_{2}$ to the current token. Unlike traditional induction heads \cite{olsson2022context} that predict just the next token of the previous occurrence of the current token, induction heads in this paper are more general. In the \nth{3} example of \cref{tab:task examples}, the token \quoteit{artist} can predict \quoteit{biro} (not \textit{the next token} of \quoteit{Frida Kahlo}) due to composition of two relations: 1) the local syntactic relation between \quoteit{Frida Kahlo} and \quoteit{biro} 2) the \texttt{OccupationOfPersons} (\textit{not} \texttt{previous-occurrence}) relation between \quoteit{artist} and \quoteit{Frida Kahlo}. 
    in our work generalize from traditional induction heads \cite{olsson2022context}. Unlike induction heads in AR that attend to (and probably output) the \textit{immediate} next token of the previous occurrence of the current token, e.g. \quoteit{A B ... A ($\rightarrow$ B)}, induction heads in GAR is in a more general form that attend to the next \textit{syntactically related} token of the previous occurrence of a token that is \textit{semantically related} (e.g. $r_{lookup}$) to the current token, e.g. \quoteit{Mary has a dog ... The girl ($\rightarrow$ dog)}, where $r_{lookup}$ = \texttt{isA}. Put another way, they are relating heads with a syntactic relation attention shift.
    
    % predict the same token which goes after the previous occurrence of the destination token and . In the prompt of \quoteit{John has car, John}, they can write information of \quoteit{car} into residual stream of the second \quoteit{John}. 
    \item \textbf{Negative-Relating Heads} attend to tokens with $\neg$\texttt{rel} semantic relations in negative generation tasks, e.g. attending to the distractive \quoteit{Mary} ($K'$) from \quoteit{John} ($Q$).
\end{itemize}

Some classes of the above heads have the so-called \textit{higher-order} counterparts. Some higher-order heads form attention relying on other basic heads of the same class. For example, True/False heads are higher-order relating heads because their queries and keys aggregate contextual information by gathering outputs from other relating attention heads. 
% \xdd{, while lower-order relating heads extract queries and keys just from intrinsic attributes of source and destination tokens (eg. apple and fruit). Note the order of attention heads is relative. Similarly, we can find higher-order induction heads, which transmit the True/False signal to higher-order local heads.}
Other higher-order heads encode higher-order relations of the statement across context. For example, induction heads capture \textit{local} induction relations in zero-shot prompts, while higher-order induction heads capture \textit{global} induction relations in few-shot prompts.  

\subsection{Circuits for More Tasks.}
\cref{tab:affirmative_g_circuits,tab:negative_g_circuits,tab:c_circuits} show the 8 circuits of Vicuna-33B we attributed for solving 8 GAR tasks (including the ones shown in \cref{fig:circuits}, $X$+ in attention patterns means the position immediately next to $X$). We observe that:
\begin{enumerate}
    \item The circuits for the same kind of tasks (e.g. affirmative generation) have the same overall structure;
    \item For some classes of heads, there are also some heads that are shared across different tasks of the same kind, e.g. 18.13 PPred. head, 15.0 Loc.$^2$ head, 16.10 Ind.$^2$ head, 7.39 Rel. head and 14.18 Rel.$^2$ (True) head for affirmative generation tasks;
    \item Pred. heads vary with different $r_{retrieve}$, because they retrieve different attributes according to the semantic relation.
\end{enumerate}
1. and 2. indicate an important mechanism of the model which is highly parameter-efficient: re-using circuit structures and attention heads as much as possible for solving similar tasks. 
\setlength{\tabcolsep}{1mm} 
\begin{table*}[!ht]
\centering
\fontsize{9}{9}\selectfont{\begin{tabular}{cccc}
\toprule
\makecell{Head Class /\\Attention Pattern} & \makecell{GendersOfPersons/=, \\ CountriesOfCities/$\in$} & \makecell{GendersOfPersons/=, \\UsagesOfThings/$\in$} & \makecell{GendersOfPersons/=, \\KindsOfThings/$\in$} \\
\midrule
Pred. / V$\rightarrow$E & 26.23 & 25.2 & 28.8, 35.8, 24.6 \\ 
PPred. / V$\rightarrow$E & 18.13, 23.9 & 19.28, 18.13, 18.1 & 18.13, 23.9, 19.28\\ 
Loc.$^2$ / Q$\rightarrow$E & 15.0, 19.28, 17.32 & 21.11, 24.27, 15.0, 17.32 & 15.0, 21.11, 14.7 \\ 
Ind.$^2$ / A$\string^$ $\rightarrow$E & 12.18, 11.1, 16.10 & 16.10, 15.4 & 16.10, 15.4 \\ 
Rel. / V$\string^$$\rightarrow$A$\string^$ & 13.31, 11.30, 11.48, 7.39 & 11.48, 13.50, 12.32, 7.39 & 13.50, 11.25, 11.2, 7.39 \\  
Rel.$^2$ / Q$\string^$ $\rightarrow$A$\string^$ & 14.18 & 14.18, 14.46 & 14.18, 14.46 \\  
Ind. / V$\rightarrow$Q & 12.8, 15.47 & 15.47, 16.33, 12.8 & 12.8, 15.47 \\  
\bottomrule
\end{tabular}
\caption{Circuits of Vicuna-33B for solving affirmative generation tasks.}
\label{tab:affirmative_g_circuits}}
\end{table*}

\setlength{\tabcolsep}{1mm} 
\begin{table*}[!ht]
\centering
\fontsize{9}{9}\selectfont{\begin{tabular}{ccc}
\toprule
\makecell{Head Class /\\Attention Pattern} & \makecell{GendersOfPersons/=, \\ CountriesOfCities/=[negate]} &  \makecell{GendersOfPersons/=, \\KindsOfThings/=[negate]} \\
\midrule
Pred. / V$'$$\rightarrow$E & 47.41, 40.21, 45.22, 37.23, 29.2, 25.8 & 46.36, 36.47, 30.19  \\ 
PPred. / V$'$$\rightarrow$E & 20.45, 20.25, 23.3, 26.8 & 23.3, 20.25, 20.45, 26.8  \\ 
R $\rightarrow$ E & 24.27 & 24.27 \\
Loc.$^2$ / Q$\rightarrow$E & 15.20, 18.9, 18.4, 17.19 & 17.19, 18.9  \\  
Rel.$^2$ / Q$\string^$ $\rightarrow$A$\string^$ & 15.51 & 15.51, 14.0\\ 
Ind.$^2$ / A$\string^$ $\rightarrow$E & 16.5, 15.4, 11.1 & 16.5, 15.4, 16.17 \\  
NRel. / K$'$ $\rightarrow$Q & 14.40 & 14.40, 9.51 \\  
Rel. / K$\rightarrow$Q & 12.50 & 8.37, 12.50, 11.48  \\  
\bottomrule
\end{tabular}
\caption{Circuits of Vicuna-33B for solving negative generation tasks.}
\label{tab:negative_g_circuits}}
\end{table*}

\setlength{\tabcolsep}{1mm} 
\begin{table*}[!ht]
\centering
\fontsize{9}{9}\selectfont{\begin{tabular}{cccc}
\toprule
\makecell{Head Class /\\Attention Pattern} & \makecell{GendersOfPersons/=, \\ CountriesOfCities/$\in$} & \makecell{GendersOfPersons/=, \\UsagesOfThings/$\in$} & \makecell{GendersOfPersons/=, \\KindsOfThings/$\in$} \\
\midrule
E$\rightarrow$E & 36.MLP & 36.MLP & 36.MLP \\ 
A$\rightarrow$E & 32.21 & \makecell{18.22, 17.10, 17.29, 15.50(A$\rightarrow$A+),\\ 25.42(A+$\rightarrow$E)} & 32.21 \\  
A$\rightarrow$A & 15.MLP &  & 15.MLP \\ 
Rel.$^2$ / Q$\rightarrow$A & 15.51, 15.30, 14.18, 14.1, 14.0 & 14.0, 14.1, 14.18 & 15.51, 14.18, 14.46 \\   
\bottomrule
\end{tabular}
\caption{Circuits of Vicuna-33B for solving classification tasks.}
\label{tab:c_circuits}}
\end{table*}

\section{E. More on Validating Attention Heads}
% \subsection{1. Illustration of Weak and Strong Intervention}

\begin{figure}[H]
\centering
\includegraphics[width=1.0\columnwidth]{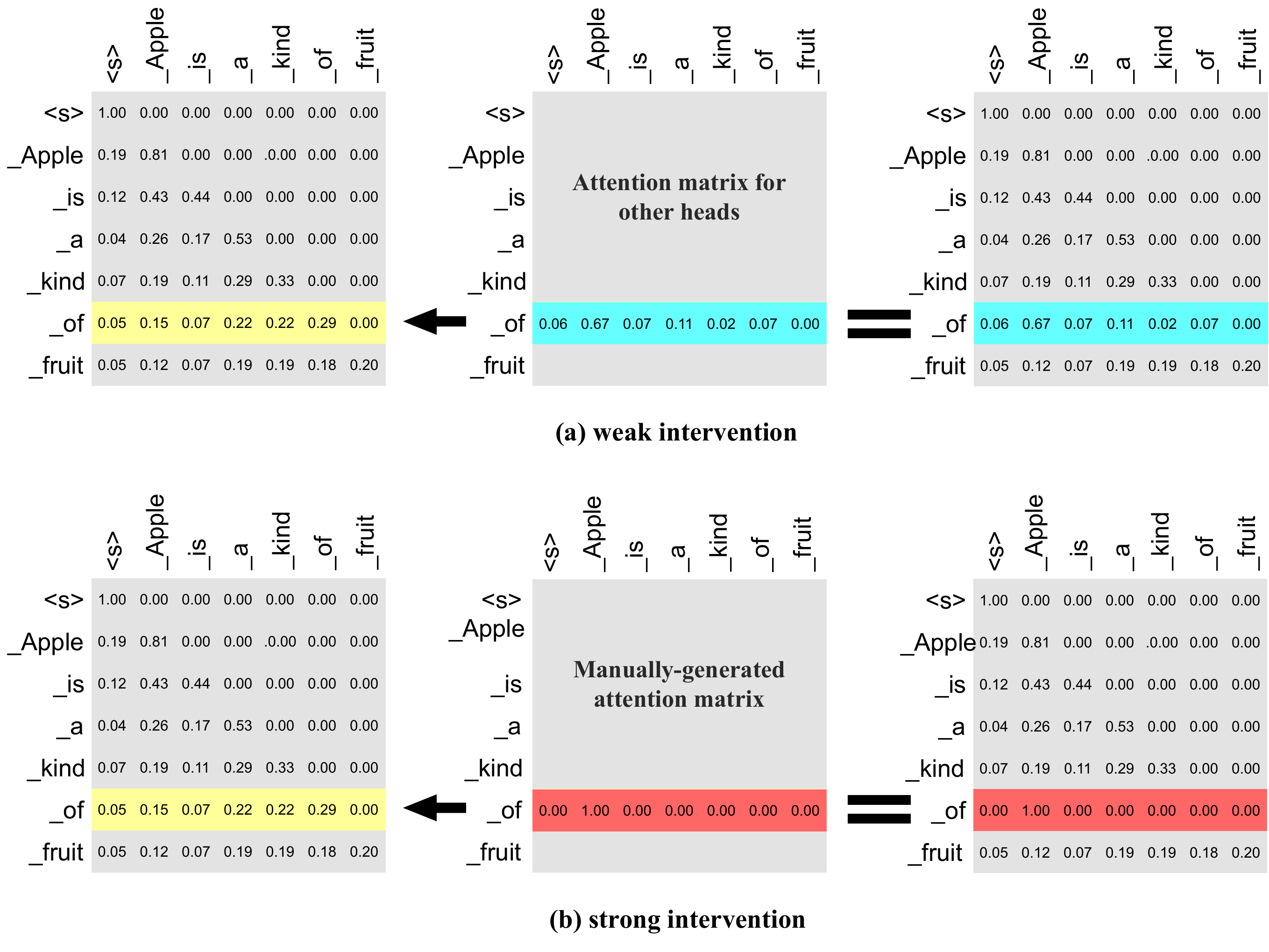}
\caption{Illustration of intervention methods.}
\label{fig:illustration_intervention}
\end{figure}

% \subsection{2. True/False Heads Discovered in Different-Sized Vicuna Models}
\setlength{\tabcolsep}{1mm} 
\begin{table}[ht]
\centering
\fontsize{9}{9}\selectfont{\begin{tabular}{ccc}
\toprule
Model &  True Heads  & False Heads  \\
\midrule
Vicuna-7B  & 11.31, 14.23 & 11.9, 9.12, 7.12      \\
Vicuna-13B  &   9.38, 9.0 &  14.35, 10.9, 10.17, 10.31    \\
Vicuna-33B  &  14.18, 14.46 &  15.51, 14.0       \\
\bottomrule
\end{tabular}
\caption{True/False heads of different-sized Vicuna models.}
\label{tab:other_true/false_head}}
\end{table}

\subsection{Inverse Intervention}
We randomly sample 500 from all 3072 classification task examples. Vicuna-33B does right on 330 of them. For these 330 examples we do inverse intervention on the True/False heads, i.e. when the answer is \textit{Yes}/\textit{No}, we apply intervention on the False/True heads while knocking out the True/False heads. As shown in \cref{tab:inverse_intervention}, a significant portion of predictions are indeed inversed with a large drop in accuracy from 100\% to 60\%.

\setlength{\tabcolsep}{1mm} 
\begin{table}[ht]
\centering
\fontsize{9}{9}\selectfont{\begin{tabular}{llll}
\toprule
                & \# Yes Predictions & \# No Predictions & Acc    \\
\midrule
orig.           & 118               & 212            & 100\%  \\
inverse interv. & 63 inversed       & 68 inversed    & 60\%   \\
\bottomrule
\end{tabular}
\caption{Inverse intervention result.}
\label{tab:inverse_intervention}}
\end{table}

\subsection{Details on Classifying Other Datasets with True/False Heads' Activation Features}
% The SNLI data comprises three elements:  premise(\textcolor{red}{P}), hypothesis(\textcolor{green}{H}), and label(\textcolor{blue}{L}), while the GoT data includes statement(\textcolor{green}{S}), label(\textcolor{blue}{L}), and others.
% In our experiments, we extracted the maximum activation values from the "hypothesis" or "statement" sections of the original data using the True/False Heads. These maximum activation values were used as new features, with the original labels retained, to train a multilayer perceptron (MLP), thereby further exploring the universality and effectiveness of these attention patterns across different datasets.

% To assess the effectiveness of these attention patterns, we designed a binary classification Multi-Layer Perceptron (MLP) model with two hidden layers. The MLP model architecture consisted of 5 input dimensions, 32 and 64 hidden units, and 2 output dimensions. \xda{TODO: try 1 or 0 hidden layer MLP}
 
\subsubsection{Data format}
Stanford Natural Language Inference (SNLI, \citet{bowman2015large}) dataset is one of the standard datasets in NLP, widely used for training and evaluating natural language inference models.
Geometry of Truth (GoT) is a dataset  of true/false statements curated by \citet{marks2023geometry} to study the structure of LLM representations of truth.
An SNLI example comprises three sections:  \textit{premise}, \textit{hypothesis}, and \textit{label}, while an GoT example includes two sections: \textit{statement} and \textit{label}.
We use the following templates to convert SNLI and GoT into question-and-answer format for testing other models (similar to evaluating LLMs on GAR, to obtain better performance we use in-context one-shot learning format which is not shown):
\begin{itemize}
  \item \textbf{SNLI}: ``Premise: \{\textit{premise}\}. Please answer with Yes or No. Can it be inferred from the premise that \{\textit{hypothesis}\}? Answer:$\rightarrow$\{\textit{label}\}''
  \item \textbf{GoT}: ``Please answer with Yes or No. Is it true that \{\textit{statement}\}? Answer:$\rightarrow$\{\textit{label}\}''
\end{itemize}
For SNLI, we replace the original labels \quoteit{entailment} and \quoteit{contradiction} with \quoteit{Yes} and \quoteit{No} respectively, and the remove examples with label \quoteit{neutral} to convert the task to one similar to GAR binary classification tasks.

\subsubsection{Feature extraction}
We use zero-shot format for extracting activation value features of True/False heads of Vicuna-33B. For SNLI, we use template: ``\{\textit{premise}\}. So \{\textit{hypothesis}\}''. For GoT, we use the same template as above for testing other models.
We extract the activation values of the True/False heads attending from the \textit{hypothesis} (SNLI) or \textit{statement} (GoT) section of the data.
We subtract the attention weights to the first token \verb|<|$s$\verb|>| from the activation values to obtain more distinctive features.

\subsubsection{Data selection and splitting}
For SNLI, to save computation, we randomly sampled 3,000 examples each from the training, validation and test set of the original dataset to use as our training, validation and test set respectively. We train 5 times with different random seed. Average accuracy with standard deviation is reported in \cref{tab:acc_mlp_true_false}.
For GoT, we use the \texttt{cities}, \texttt{sp\_en\_trans}, \texttt{companies\_true\_false} and \texttt{counterfact\_true\_false} subsets among the 12 subsets in Table 1 of \citet{marks2023geometry}.
We don't use \texttt{larger\_than} and \texttt{smaller\_than} because the judgement of their truthfulness are attributed to the function of MLPs rather than attention heads \cite{hanna2024does}. We don't use \texttt{cities\_cities\_conj}, \texttt{cities\_cities\_disj}, \texttt{neg\_cities} and \texttt{neg\_sp\_en\_trans} because they require logical operations which are beyond the scope of MI analysis of this work.
We also exclude \texttt{common\_claim\_true\_false} and \texttt{likely} because they are either too difficult or too vague for even the full Vicuna-33B model to achieve reasonable accuracy, let alone the activation features of its True/False heads.
For each selected subset, we use 5-fold cross-validation and for each fold we further split the training set into training and validation sets with 3:1 ratio. We report the average accuracy with standard deviation for each subset in \cref{tab:acc_mlp_got_subsets}. The average result of the 4 subsets is in shown \cref{tab:acc_mlp_true_false}.

\subsubsection{Hyper-parameters}
For all models, we train with Adam optimizer without weight decay for 30 epochs, and save the model with the highest validation accuracy to evaluate on test set.
The hyper-parameters of the MLP model architecture and training are in \cref{tab:hparam_training_mlp}. We base our code on Huggingface Transformers 4.36.0 and PyTorch 2.3.1. All experiments are run on a workstation with one A800 GPU with 80GB HBM.

\setlength{\tabcolsep}{1mm} 
\begin{table}[ht]
\centering
\fontsize{10}{10}\selectfont{\begin{tabular}{lll}
\toprule
Hyper-parameter  & value    \\
\midrule
MLP input dim.  & 4      \\
MLP hidden dim.  & 32      \\
MLP hidden activation  & ReLU      \\
MLP output dim.  & 2      \\
learning rate  & 1e-3 \\
batch size  & 32 \\
Adam ($\beta_1$, $\beta_2$) & (0.9, 0.999) \\
Adam $\epsilon$  & 1e-8 \\
\bottomrule
\end{tabular}
\caption{Hyper-parameters of training MLP classifiers with activation values of True/False heads of Vicuna-33B as features.}
\label{tab:hparam_training_mlp}}
\end{table}

\setlength{\tabcolsep}{1mm} 
\begin{table}[ht]
\centering
\fontsize{9}{9}\selectfont{\begin{tabular}{lllll}
\toprule
Model  & \makecell[l]{cities\\acc (\%)} & \makecell[l]{sp\_en\\acc (\%)} & \makecell[l]{companies\\acc (\%)} & \makecell[l]{counterfact\\acc (\%)}  \\
\midrule
Random Guess & 50  & 50  & 50 &  50 \\
GPT-2-medium  & 51.27 & 50.85 & 51.58  & 50.47   \\
GPT-2-XL  & 50.53 & 47.46 & 52.75   &  50.67 \\
\B{MLP}  & \B{84.29$\pm$1.0} & \B{96.61$\pm$2.6} & \B{85.33$\pm$3.0} & \B{76.27$\pm$0.7}    \\
Vicuna-7B  & 80.35 & 97.18 & 88.42 &  72.97  \\
Vicuna-33B & 89.97 & 98.02 & 90.08 & 80.70 \\
\bottomrule
\end{tabular}
\caption{Accuracy of different models and MLP classifiers with activation values of True/False heads of Vicuna-33B as features on different subsets of GoT.}
\label{tab:acc_mlp_got_subsets}}
\end{table}
% \end{comment}
\end{document}